\pdfoutput=1

\documentclass[11pt]{article}

\usepackage{comment}

\usepackage[preprint]{acl}

\usepackage{times}
\usepackage{latexsym}

\usepackage[T1]{fontenc}

\usepackage[utf8]{inputenc}

\usepackage{microtype}

\usepackage{inconsolata}

\newcommand{\n}{\newline}  
\def\t+{+}

\usepackage{graphicx}
\usepackage{subcaption} 
\usepackage{listings}
\usepackage{xcolor}  

\lstdefinelanguage{json}{
    basicstyle=\ttfamily\small,
    numbers=right,
    numberstyle=\scriptsize,
    stepnumber=1,
    showstringspaces=false,
    breaklines=true,
    frame=single,
    backgroundcolor=\color{gray!5},
    literate=
     *{:}{{{\color{black}{:}}}}{1}
      {,}{{{\color{black}{,}}}}{1}
      {\{}{{{\color{black}{\{}}}}{1}
      {\}}{{{\color{black}{\}}}}}{1}
      {[}{{{\color{black}{[}}}}{1}
      {]}{{{\color{black}{]}}}}{1},
}

\title{\textit{PromptTailor}: Multi-turn Intent-Aligned Prompt Synthesis for\\ Lightweight LLMs}

\author{Yizhou Xu \and Janet Davis \\
  Whitman College, Department of Computer Science \\
  \texttt{\{xuy2, davisj\}@whitman.edu}}
  
\begin{document}
\maketitle

\begin{abstract}

Lightweight language models remain attractive for on-device and privacy-sensitive applications, but their responses are highly sensitive to prompt quality. For open-ended generation, non-expert users often lack the knowledge or time to consistently craft high-quality prompts, leading them to rely on prompt optimization tools. However, a key challenge is ensuring the optimized prompts genuinely align with users’ original intents and preferences. We introduce \textit{PromptTailor}, a system for \textbf{controllable prompt generation for open-ended text} that \textbf{improves model output quality by intent-aligned prompt synthesis}. \textit{PromptTailor} expands minimal user instructions into rich, domain-aware prompts while preserving the user’s stated preferences. The system is a quantized Llama3-8B model fine-tuned with a lightweight LoRA adapter on 12{,}300 prompt-refinement dialogues spanning 41 everyday domains, distilled from three stronger LLMs. The adapter attaches to any Llama3-8B base, enabling edge deployment. In human and LLM-judge evaluations across multiple target models and optimization baselines, \textit{PromptTailor} yields higher preference rates than chain-of-thought prompting and matches or surpasses state-of-the-art prompt optimization methods while requiring fewer model calls (e.g., 3 vs.\ 9). These results show that a compact student, guided by powerful teachers, can learn effective prompt-generation strategies that enhance response quality while maintaining alignment with user intent.

\end{abstract}

\begin{figure*}[h]
  \centering
  \includegraphics[width=\linewidth]{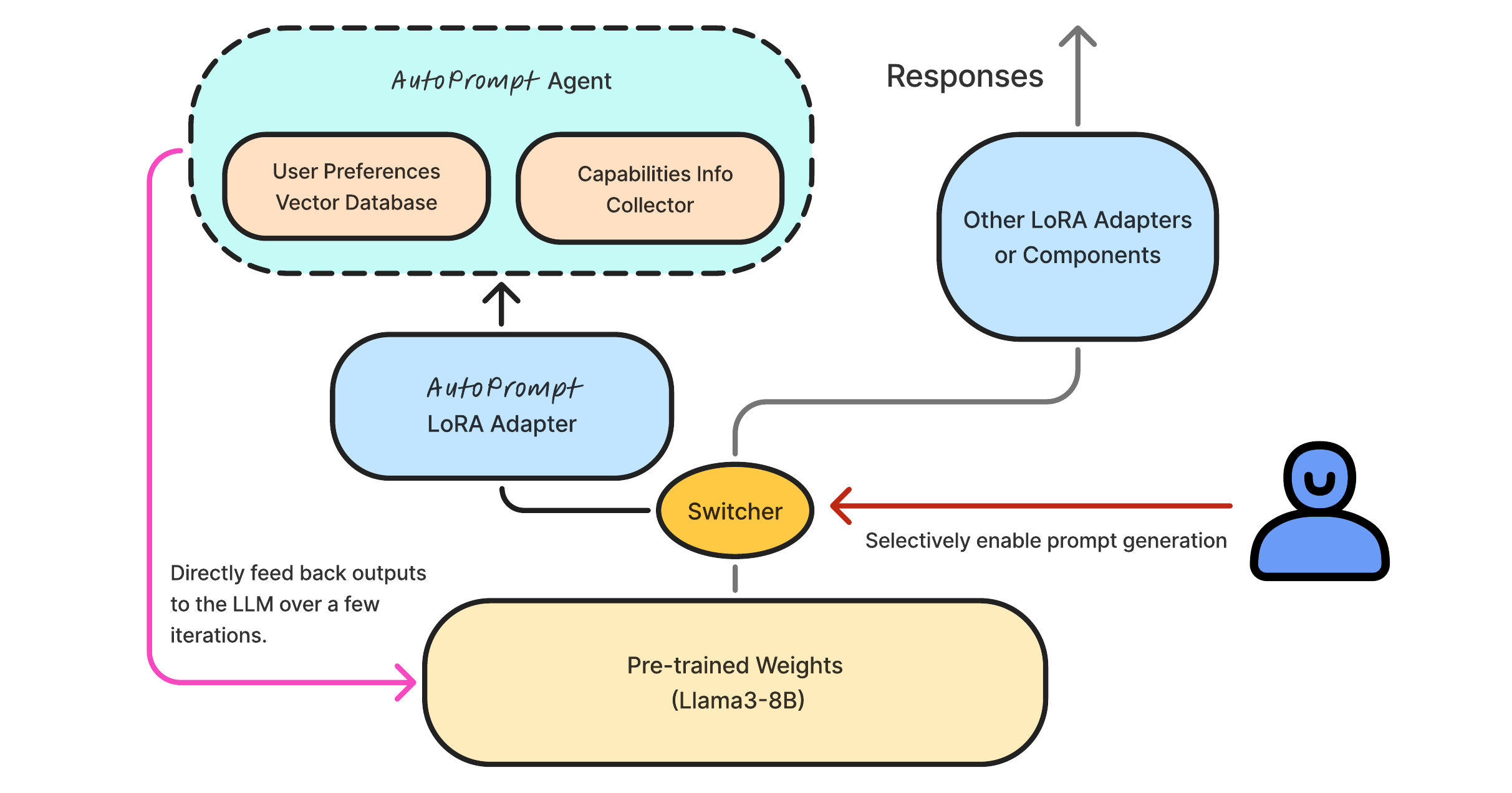}
  \caption{\label{fig:agent}\textbf{An example of integrating \textit{PromptTailor} into an existing Llama3-8B model.} While multiple integration strategies are possible, our proposed structure allows the user to choose whether to activate the prompt generation capability. The intermediate output is then fed directly back into the model to produce results. The \textit{PromptTailor} operates as a separate LoRA adapter agent, ensuring that it does not impact other LoRA adapters or components within the model.}
\end{figure*}

\section{Introduction}

We have seen the superb performance of recent reasoning models like OpenAI o3 \citep{pfister2025o3notagi}. Those language models are capable of generating high-quality responses without requiring several iterations for prompt optimization. However, the lightweight language models are still useful due to their ability to run on mobile or personal devices. For those models with less than 8B parameters and especially for quantized ones, the quality of the prompt still has a high impact on the response \citep{liu2023pretrain}. It might be difficult and time-consuming for non-expert users to self-design the prompt for the large language models (LLMs). Non-expert users often lack the experience or domain knowledge necessary to craft high-quality prompts and may instead rely on prompt optimization tools. However, existing prompt optimization strategies are sometimes difficult for general users to use or interpret. This also raises the question of whether the optimized prompts truly align with the users’ original intent and preferences.

In this paper, we introduce \textit{PromptTailor}\footnote{Link for the accompanying generation prompts/templates: \url{https://github.com/riverside234/data-template}.}, an LLM-based prompt generation system that transforms user intents and preferences into enriched and elaborated prompts, while maintaining alignment with the original intent. The system is capable of generating elaborate prompts that most prior methods cannot produce, and these prompts are human-readable, making them user-friendly and easy to edit. Specifically, we fine-tuned a quantized 4-bit version of Llama3-8B \citep{grattafiori2024llama} using a novel synthetic dataset. To ensure the lightweight LLM acquires sufficient prompt generation and optimization capabilities, we leverage three efficient yet powerful LLMs to construct our synthetic dataset. This comprehensive dataset spans over 41 diverse domains, featuring detailed prompt generation and optimization dialogues across its 12,300 instances. 

In our synthetic dataset formulation, we introduce a crucial mechanism: \textbf{"capabilities."} This mechanism ensures that the generated prompt remains intent-aligned while still providing the LLM with room to improvise during prompt generation. We instruct the LLM to gather capabilities information from two sources: (1) a list of capabilities deemed necessary to complete the task, and (2) capabilities explicitly mentioned in or inferred from the user's original intent. In subsequent turns, the model is instructed to balance these derived capabilities during prompt optimization.

Since our \textit{PromptTailor} is essentially based on a fine-tuned LoRA adapter, it can be seamlessly integrated into an existing LLM. This integration allows for easy usage by non-expert users and also preserves system efficiency. There are various methods for integrating our system. See Figure~\ref{fig:agent} for an example of the system structure. In our proposed structure, \textit{PromptTailor} acts as a separate LoRA agent that does not affect the functionality of other components in the existing LLM. Users can selectively activate the agent when prompt generation capabilities are required. Our agent runs automatically for several turns to generate an optimized prompt. To further enhance generation quality, the agent system incorporates two additional components: a local user preference database and a custom-designed capabilities information collector (see Figure~\ref{fig:flow} for details).  

For evaluation, we conduct both human and automated assessments to record performance improvements on Lightweight LLMs using different prompting methods. For automated assessment, we designed an LLM-based evaluator that conducts pairwise comparisons on the quality of open-ended responses from LLMs. We provide three pieces of information during evaluation: (1) user original intent and preferences, (2) the response generated by the original user intent and preferences, and (3) the response generated by the LLM using a prompt produced by our \textit{PromptTailor} (or another prompting method). 

The LLM evaluator or human participants are asked to provide scores (1-10) based on two metrics: (1) how well the response aligns with the user's original intent and preferences, and (2) overall response quality, including correctness, usefulness of information, and a well-structured style, etc. The two scores are averaged and compared with the other method. The method with the higher average score is considered the "winner" or "better." Our automated evaluation results indicated that our \textit{PromptTailor} outperforms CoT prompting and is equivalent to recent prompt optimization techniques such as Evoke \citep{hu2023evoke}. However, our method is more efficient, requiring 
three model calls compared to Evoke's nine calls.

Additionally, we assessed the effectiveness of prompts generated by \textit{PromptTailor} across different LLMs, including those with capabilities stronger or weaker than our base model. The findings from both human and automated evaluation suggest that a more substantial improvement can be seen on weaker models, while a noticeable improvement can still be observed on stronger models.

\section{Related Works}
Recent approaches aim to improve LLM performance by guiding their behavior through structured prompting strategies. Multi-expert Prompting \citep{long-etal-2024-multi-expert} steers an LLM to simulate and aggregate several expert viewpoints, then selects the safest, most factual, and useful response. ExpertPrompting \citep{xu2023expertprompting} auto-generates task-specific expert personas that steer LLMs to deliver more accurate, expert-level answers. The Evoke \citep{hu2023evoke} reframes prompt engineering as a reviewer–author dialogue in which two instances of the same LLM iteratively critique and rewrite a prompt—and, aided by a selector that feeds only “hard” examples—converges on a refined instruction in just three cycles (nine API calls). However, some of these approaches assume the existence of a well-structured short prompt to optimize. In contrast, our approach generates and refines a prompt directly from user intent and preferences. In addition, we shrink the overall framework and system by distilling knowledge into a lightweight model that can run on edge devices. 

Adapting an LLM to a specific domain or task commonly involves fine-tuning on in-domain data \citep{lu2024fine}. We instead treat prompt refinement itself as a specialization mechanism. Because no public dataset for prompt generation and optimization exists, we leverage three powerful LLMs to synthesize training data. To assess the quality of the optimized prompts generated by our fine-tuned model, we build an LLM-based evaluator. Existing LLM-based evaluators such as GPTScore \citep{fu2024gptscore} repurposes GPT models as instruction-guided, training-free evaluators that flexibly score generated text on any user-specified criteria. CritiqueLLM \cite{ke2023critiquellm} trains an open-source evaluator to generate both scalar scores and textual critiques using supervision distilled from GPT-4. The model achieves GPT-4-level alignment with human preferences while providing actionable feedback for response improvement. G-Eval \citep{liu-etal-2023-g} couples GPT-4’s chain-of-thought reasoning with rubric-style form filling to score NLG outputs in stronger alignment with human judgments. However, we noticed that absolute score assignments by LLMs can be arbitrary and less reliable when conducting cross-data comparisons. To address this, we use pairwise comparisons instead of absolute scoring to evaluate the quality of responses generated from optimized prompts.

\section{Methods}

\subsection{Synthetic Data Generation}

We built a synthetic dataset of high-quality prompt generation and optimization dialogues for fine-tuning. The dataset covers 41 diverse domains (see Appendix~\ref{sec:domains}), from legal advice and business inquiries to everyday tasks such as cooking. To keep the fine-tuned model from merely mimicking the style and strength of a single system, we utilized three distinct yet cost-efficient LLMs—GPT-4o-mini \citep{openai2024gpt4o-mini}, Claude 3 Haiku \citep{anthropic2024claude3haiku}, and Grok-3-mini \citep{xai2025grok3mini}. We generated the majority of dialogues (8,200) using Grok-3-mini due to its superior performance, with the remaining instances contributed by the other two models, yielding a total of 12,300 dialogues.

In each dialogue, the LLM first simulates a user by generating a domain-specific intent and a set of preferences. We then use the LLM to generate and refine a prompt based on this intent and preference. To clearly identify the user's goals, the LLM is instructed to elaborate on the purpose, context, and desired outcomes derived from the intent and preference. The model is later asked to strictly adhere to this information when generating and refining the prompt.

While testing our framework, we observed that directly generating prompts from user intent sometimes led to the LLM misinterpreting the term "prompt". In some cases, the model generated "prompts" intended for academic reports or assignments, misaligning with the objective of eliciting specific LLM behavior. To address this, instead of generating a prompt directly, we instruct the LLM to first craft a plan for an LLM agent based on the user’s intent and preferences, and then generate a corresponding prompt for that agent.

A common challenge we identified is that while generated prompts should ideally align with user intent, prompts that perfectly align might not always be optimal for eliciting the best LLM response. To address this trade-off, we introduce \textbf{"capabilities"} as a crucial mechanism. This ensures that the generated prompt remains intent-aligned while still providing the LLM with room to improvise and optimize. For example, in the context of a financial report generator, a relevant capability might be "identifying and explaining market trends." We instruct the LLM to gather capability information from two distinct sources: (1) a list of capabilities deemed necessary to complete the task, and (2) capabilities explicitly mentioned in or inferred from the user's original intent. This aggregated capability information is then passed to the subsequent turns, and we instruct the model to balance between the capabilities derived from these two sources when formulating its response.

\textbf{Each dialogue in our dataset contains the following four turns (See Appendix~\ref{sec:sydata}):}

\begin{itemize} 
    \item \textbf{User Intent and Preference:} To ensure \textbf{generalization to real-world prompts}, we first generated realistic user intents designed to mimic typical user inputs (e.g., “I want to draft an overview of the safety precautions and travel advisories...” ). To diversify the dataset, some intents were made more detailed, while others were intentionally underspecified. Detailed intents typically consist of a short paragraph with specific user directives, whereas underspecified ones are one or two sentences that provide only broad guidance. Additionally, extra user preference information related to the context—such as “the user prefers short, concise responses” or “the user enjoys cooking”—was generated and appended to the end.
    
    \item \textbf{Intent Understanding and Base Prompt Formulation:} In this stage, the LLM is instructed to perform the following steps: (1) Intent and Preference Categorization: Convert raw user intents and preferences into two distinct categories: "purpose" and "context." (2) Desired Outcome Analysis: Carefully analyze the desired outcome (or output) the user expects from the LLM agent. (3) Capability Information Gathering: \textit{Explicit/Inferred Capabilities:} Extract any capabilities explicitly mentioned in or inferred from the user's original intent; \textit{Task-Required Capabilities}: Assess the type of LLM agent required for the user's task, and identify its essential capabilities (4) Agent Plan Generation: Provide a short, high-level plan for the agent. (5) Initial Prompt Generation: Generate a simple, foundational prompt for the agent.

    \item \textbf{Prompt Optimization Suggestions:} In this stage, the LLM is instructed to first carefully review and briefly summarize the information passed from the previous turn, including the purpose, context, and desired outcome. Subsequently, the model rebalances the capability information from the last turn to generate an optimized list of capabilities. 
    Finally, it produces five concrete suggestions for optimizing the original agent prompt: 
    (1) define an appropriate role for the agent; 
    (2) use precise, domain-specific terminology; 
    (3) provide any necessary extra background context; 
    (4) add missing details for completeness; and 
    (5) specify the desired output format, length, and formatting.
    \item \textbf{Final Prompt Generation:} We asked the LLM to generate a optimized prompt based on the plan and suggestions they provided. 
\end{itemize}

We conducted an additional data filtering step using Grok-3-mini to discard low-quality refined prompts, generating additional prompts to replace each discarded one. Finally, we randomly selected ten generated dialogues from each domain (41 domains) to form a test set of 410 dialogues, with the remaining dialogues serving as the training set. We will use this test set in subsequent experiments.

\begin{figure*}[h]
  \centering
  \includegraphics[width=\linewidth]{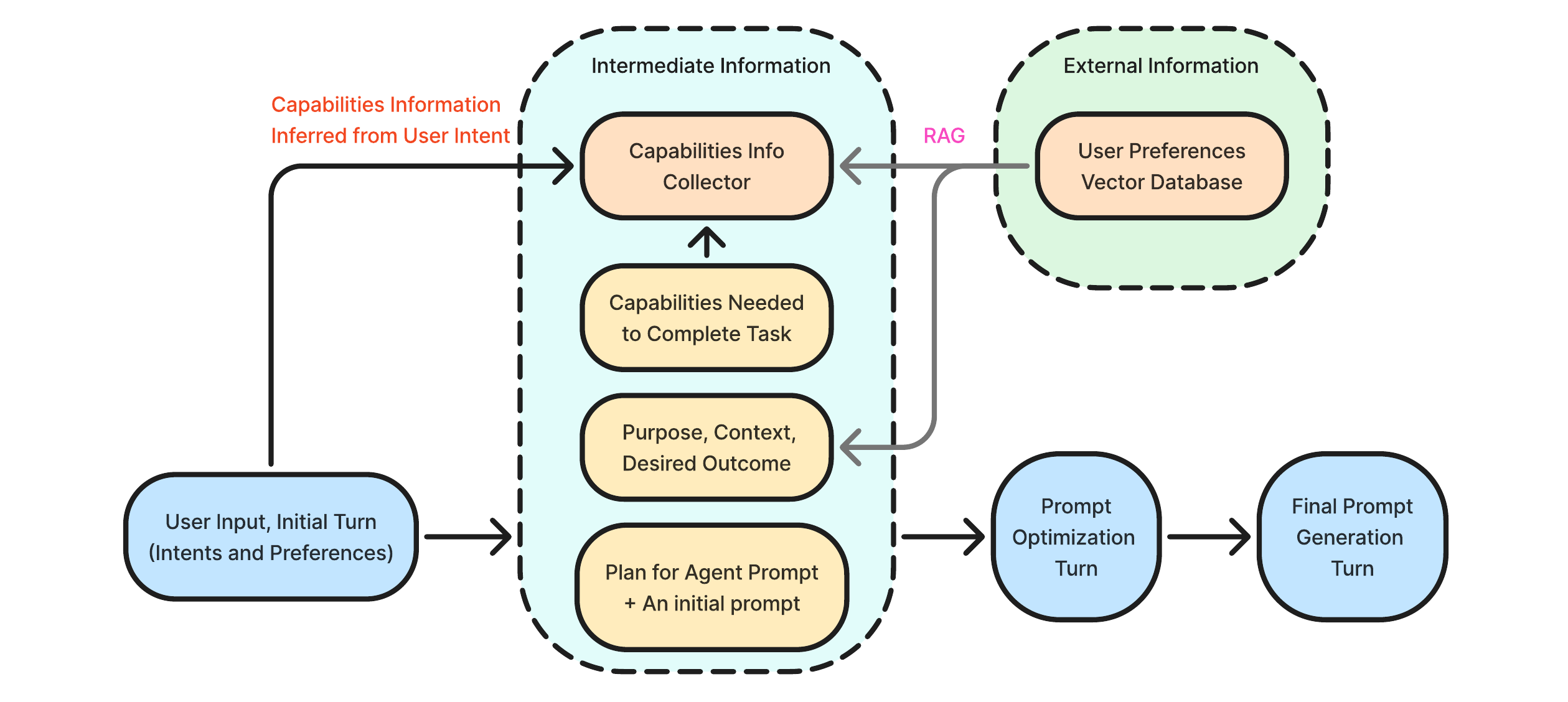}
  \caption{\label{fig:flow}\textbf{Agent System Structure:} The agent processes initial user input through three consecutive model calls. The agent's dialogue flow mirrors that of our synthetic data. The system utilizes RAG to retrieve historical user preference information from a local preference database. Additionally, capabilities information is collected from various sources by a dedicated collector component to help the LLM adjust its responses accordingly in subsequent turns.}
\end{figure*}

\subsection{Model Fine-Tuning}
We fine-tuned a quantized 4-bit version of Llama3-8B \citep{unsloth2024Llama3} on our synthetic dataset. Quantization reduces the memory footprint, thereby enabling fine-tuning on a single NVIDIA A100 GPU and inference on resource-constrained devices. 

\paragraph{Training Details}
Each dialogue is first converted into the Llama 3 chat format with the required special tokens. We employed the Hugging Face TRL \texttt{SFTTrainer} with LoRA (Low-Rank Adaptation) for parameter-efficient fine-tuning \citep{hu2021lora}. Guided by prior work \citep{halfon2024stay} on Llama3-8B fine-tuning, we adopted a total batch size of 32 (8 per device, using $4\times$ gradient accumulation) and the 8-bit AdamW optimizer. We set the learning rate to $4\times10^{-5}$ and used a \texttt{constant\_with\_warmup} scheduler, with a warmup ratio of 0.05. For LoRA, we used a rank of 32, $\alpha = 64$, and a dropout rate of 0.05.

To determine an appropriate stopping point, we ran some exploratory experiments for three epochs. The training losses plateaued at approximately 0.5 during the second epoch. Consequently, our final experiment trained on the full dataset for two epochs, yielding a training loss of 0.537.

\subsection{Agent System Structure}
Most previous prompt optimization methods require external programs and additional computation to revise prompts; however, our system can be integrated directly into the base model, preserving efficiency. See Figure~\ref{fig:agent} for an integration example. Our \textit{PromptTailor} is essentially a fine-tuned LoRA adapter for the Llama3-8B model. During inference, the integrated model can selectively activate this LoRA adapter, enabling dual capabilities: generating prompts or performing other tasks. The approach is compatible with LLMs other than the Llama3-8B model, as long as the new model is fine-tuned on our synthetic dataset. 

In our proposed agent system, after the user provides an initial intent and preferences for a desired prompt, the agent automatically executes three turns to produce a final optimized prompt. (See Figure~\ref{fig:flow} for more details.) We designed the prompt generation flow of the agent system to match the structure of the synthetic data on which it was fine-tuned. To maintain efficiency across these turns, we employ a context window optimization strategy. Instead of passing the entire conversational history to the LLM in each step, we selectively curate and structure prior information into a concise intermediate format.

The system utilizes Retrieval-Augmented Generation (RAG) to retrieve historical user preference information from a local database. Following retrieval, the "purpose and context" section is updated with this additional user preference information. Furthermore, capabilities information is systematically collected from various sources by a dedicated collector component, enabling the LLM to adjust its responses accordingly in subsequent turns.

We identify three distinct channels through which capabilities information can be collected. First, users may explicitly include capability details within their initial intent and preferences, which the LLM is instructed to infer and extend based on the input. Second, if a user has previously conducted related tasks, relevant capabilities information can be retrieved from the database using RAG. Lastly, a list of capabilities deemed necessary to complete the task is generated by the LLM during the first turn. The model is then instructed to balance the capabilities derived from these three sources in subsequent turns.

\begin{table*}[h]
  \centering
  \begin{tabular}{lcccc}
    \hline
    \textbf{Model} & \textbf{Response Comparison} & \textbf{1 (First Better)} & \textbf{2 (Second Better)} & \textbf{0 (Same)} \\
    \hline
    \textbf{Llama3-8B} & \textbf{Original vs. CoT} & 177 & 216 & 17 \\
    \textbf{Llama3-8B} & \textbf{Original vs. Expert} & 170 & 225 & 15 \\
    \textbf{Llama3-8B} & \textbf{Original vs. Evoke(Llama3)} & 145 & 253 & 12 \\
    \textbf{Llama3-8B} & \textbf{Original vs. Llama3(FT)}       & 141 & 248 & 21 \\
    
    \hline
  \end{tabular}
  \caption{\label{tab:auto1}
    \textbf{Results of automated assessments, sample size 410}. This table compares responses from Llama3-8B generated using prompts from the original user intent versus various optimization methods. “Original” refers to the original user intents, “CoT” refers to CoT prompting, "Expert" refers to ExpertPrompting, "Evoke" refers to the Evoke prompting framework that uses Llama3-8B, and "Llama3(FT)" refers to using our fine-tuned \textit{PromptTailor} model to optimize the prompt. Each cell shows the number of judgments favoring the response from the original prompt (1), the optimized prompt (2), or indicating no difference (0).
  }
\end{table*}

\begin{figure*}[h]
  \centering
  \includegraphics[width=\linewidth]{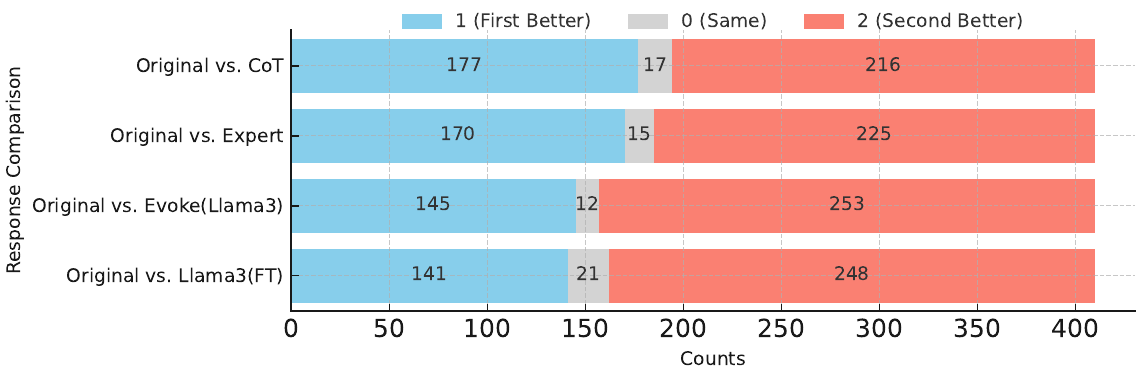}
  \caption{\label{fig:auto1}\textbf{The bar chart version of the table above}. Each cell shows the number of judgments favoring the response from the raw user intent and preference (1), the optimized prompt (2), or indicating no difference (0).}
\end{figure*}

\begin{table*}
  \centering
  \begin{tabular}{lcccc}
    \hline
    \textbf{Model} & \textbf{Response Comparison} & \textbf{1 (First Better)} & \textbf{2 (Second Better)} & \textbf{0 (Same)} \\
    \hline
    \textbf{Llama3-8B} & \textbf{Original vs. Llama3(FT)}       & 141 & 248 & 21 \\
    \hline
    \textbf{Gemma2-9B} & \textbf{Original vs. Llama3(FT)}         & 147 & 230 & 33 \\
    \hline
    \textbf{Mistral-7B} & \textbf{Original vs. Llama3(FT)}        & 103 & 298 & 9 \\
    \hline
    \textbf{Llama3-70B} & \textbf{Original vs. Llama3(FT)}     & 155 & 232 & 23 \\
    
    \hline
  \end{tabular}
  \caption{\label{tab:auto2}
    \textbf{Results of automated assessments, sample size 410}. We conduct a comparative analysis of responses from four LLMs (all quantized versions), evaluating their performance when provided with the original user intent (Original) versus prompts optimized by our fine-tuned optimizer \textit{PromptTailor}. “Original” refers to the original user intents and "Llama3(FT)" denotes using our fine-tuned Llama3 model (\textit{PromptTailor}) to refine the prompt. Each cell shows judgments favoring the response from the original prompt (1), the optimized prompt (2), or same (0).
  }
\end{table*}

\begin{figure*}[h]
  \centering
  \includegraphics[width=\linewidth]{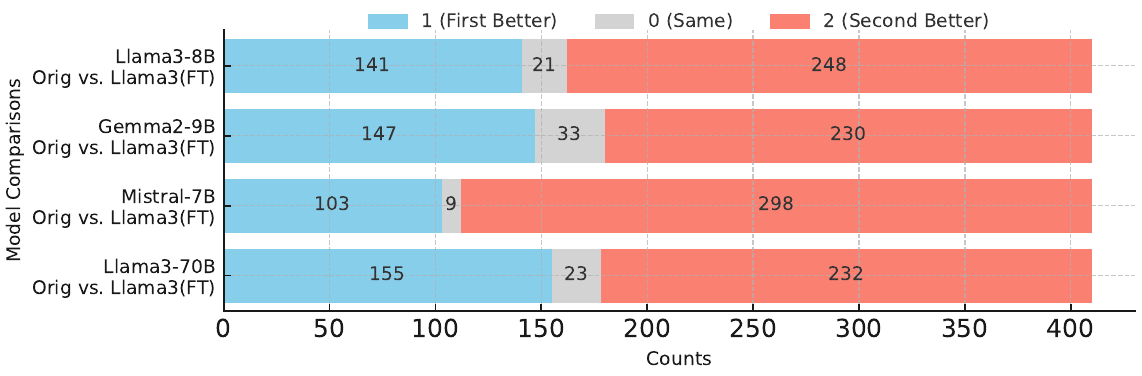}
  \caption{\label{fig:auto2}\textbf{The bar chart version of the table above}. Each cell shows the number of judgments favoring the response from the raw user intent and preference (1), the optimized prompt (2), or indicating no difference (0).}
\end{figure*}

\section{Experiment Results and Discussion}

\subsection{Automated Assessments}
To evaluate open-ended generations from language models, we designed an LLM-based evaluator rather than relying on standardized benchmarks. We selected a reasoning model, OpenAI o1-mini \citep{openai2024o1} as the backbone model for the evaluator, balancing cost and efficiency given the large volume of text to assess. The evaluation focuses on comparing responses generated from original user intents with those from optimized prompts across diverse methods and LLMs. However, score assignment by LLMs can be arbitrary and less reliable for cross-dataset comparisons. To address this, we use pairwise comparisons instead of absolute scoring to evaluate response quality.

To evaluate responses, our LLM-based system is provided with raw user intents and preferences along with two candidate responses. The evaluation scoring is based on two criteria: (1) the degree to which the response aligns with the original user intent, and (2) the overall quality of the response. The system first compares these responses by analyzing each and assigning two integer scores (1-10) based on these criteria. The two scores are then averaged and compared with those from the other method. To determine a winner, the LLM is instructed to return an integer: 0 (indicating equal quality), 1 (first response is better), or 2 (second response is better), based on the assigned scores. This evaluation is repeated five times, and the final winner is selected based on the mode. In cases with multiple modes, the experiment runs for an additional round, although this rarely occurs.

\paragraph{Implementation details} Our test set comprises 410 raw user intents and preferences. These are then fed into \textit{PromptTailor}, which first analyzes user intents to suggest prompt components, and subsequently generates a complete prompt using a total of three model calls. We conduct pairwise comparisons to evaluate the responses of LLMs, considering prompts generated by both \textit{PromptTailor} and other methods. In our experiments, external information sources, such as a local user preference database, were not included.

\begin{table*}[h]
  \centering
  \begin{tabular}{lcccc}
    \hline
    \textbf{Model} & \textbf{Response Comparison} & \textbf{1 (F. Better)} & \textbf{2 (S. Better)} & \textbf{0 (Same)} \\
    \hline
    \textbf{Llama3-8B} & \textbf{"Without Cap" vs. "Cap" (Llama3-8B)} & 171 & 212 & 27 \\
    \textbf{Llama3-8B} & \textbf{"Without Cap" vs. "Cap" (Gemma2-9B)} & 152 & 218 & 40 \\
    
    \hline
  \end{tabular}
  \caption{\label{tab:auto3}
    \textbf{Results of our ablation study, evaluated on responses from a base Llama3-8B model, sample size 410.} The prompts for this base model were generated by two separate, fine-tuned optimizers: one built on Llama3-8B and the other on Gemma2-9B. The table compares the effectiveness of each optimizer when trained "Cap" (our full method) versus "Without Cap" (without the capabilities mechanism). Each cell shows the number of judgments favoring the response from "Without Cap" (1), "Cap" (2), or indicating no difference (0).
  }
\end{table*}

\begin{figure*}
  \centering
  \includegraphics[width=\linewidth]{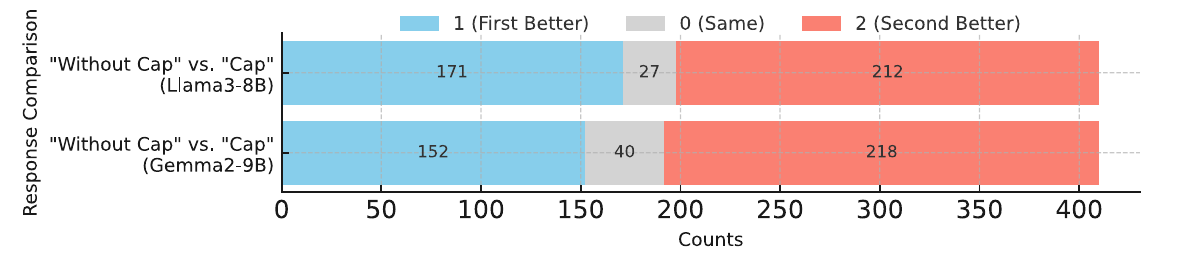}
  \caption{\label{fig:auto3}\textbf{The bar chart version of the table above}. Each cell shows the number of judgments favoring the response from "Without Cap" (1), "Cap" (2), or indicating no difference (0).}
\end{figure*}

\paragraph{Baselines}
We compare \textit{PromptTailor} with the following techniques and baselines:
(1) \textbf{Original} refers to using the unmodified Llama3-8B model to transform raw user intents and preferences into a short prompt (a few sentences). This transformation is necessary because the raw intents may contain meta-text (e.g., “Show me a prompt...”) or preference information that influences the focus of the response, such as “the user is interested in swimming.”
(2) \textbf{CoT} \citep{wei2022chain} refers to chain-of-thought prompting. 
(3) \textbf{Expert} refers to ExpertPrompting \citep{xu2023expertprompting}, a method that steers LLMs to adopt specialized expert roles and coordinate those roles.
(4) \textbf{Evoke (Llama3)} refers to applying the Evoke prompt refinement framework \citep{juneja2024taskfacet} to edit the prompt. For a fair comparison, we replace the original GPT-4-based editor with the unmodified Llama3-8B model. The framework involves three rounds of editing, comprising a total of nine calls: three Author, three Reviewer, and three Selector steps.

The results of the pairwise comparisons between baselines and our \textit{PromptTailor} are presented in Table~\ref{tab:auto1} and Figure~\ref{fig:auto1}. Compared to CoT prompting, our fine-tuned model achieves superior performance, with 248 favorable outcomes. In terms of performance, Evoke is largely on par with our framework, showing comparable favorable outcomes against original prompts. However, \textit{PromptTailor} distinguishes itself by requiring significantly fewer model calls—three compared to Evoke's nine—making it a more efficient and preferable solution for edge device deployment.

\paragraph{Cross Model Comparisons} We conduct a cross-model comparison, comparing responses generated by four distinct quantized LLMs \citep{unsloth2024gemma2, unsloth2024llama3_70b, unsloth2024mistral} using both raw user intents (transformed by the unmodified Llama3-8B) and optimized prompts produced by \textit{PromptTailor}. As shown in Table~\ref{tab:auto2} and Figure~\ref{fig:auto2}, smaller models such as a quantized 4-bit version of Mistral-7B benefit significantly from optimized prompts, with 298 favorable judgments for optimized prompts versus 103 for original ones. \textbf{Even with a much stronger target model—Llama-3 70B, whose base capability exceeds that of the 8B optimizer—prompt optimization remains beneficial.} The result demonstrates greater gains for weaker models, with some improvement also seen for stronger ones.

\subsection{Analysis on the "Capabilities" Mechanism}

To isolate the impact of our proposed "capabilities" mechanism, we conducted an ablation study. We compared two models fine-tuned on distinct versions of our synthetic dataset: (1) \textbf{"Without Cap"}, trained on an earlier version of the dataset lacking the capabilities framework, and (2) \textbf{"Cap"}, trained on the final dataset incorporating this mechanism. During inference, the \textbf{"Cap"} model maintains context across its three optimization turns by selectively curating and structuring information from prior steps—a process we refer to as context window optimization (Figure~\ref{fig:flow}). In contrast, the \textbf{"Without Cap"} model performs a direct three-turn optimization without this guidance. To ensure our findings generalize, we fine-tuned quantized 4-bit versions of Llama3-8B and Gemma2-9B to serve as two separate prompt optimizers. During the evaluation, prompts from each of these optimizers were fed to a base Llama3-8B model. We then used our LLM-as-a-Judge to evaluate the final responses generated by this base model. As shown in Table~\ref{tab:auto3}, for both the Llama3-8B and Gemma2-9B optimizers, prompts generated using the With-Cap method were generally preferred, confirming the efficacy of the capabilities mechanism.

\subsection{Human Assessments}

To complement our automated evaluation, we conducted a small human study with 10 participants. The results, detailed in Appendix~\ref{sec:humana}, generally align with the findings from our automated assessments. Participants showed a clear preference for responses generated from optimized prompts over those from the original, unrefined user intents. However, we observed some variance in participant sensitivity; while some evaluators identified significant nuances in response quality, others perceived the outputs as being broadly similar. A larger-scale human evaluation to investigate these subtleties is left as future work.

\section{Conclusion and Future Work}

This paper introduces \textit{PromptTailor}, an agent designed to automatically generate refined prompts that align with user intents and preferences. We fine-tuned a lightweight Llama3-8B model using synthetic data derived from three powerful large language models. This approach produces improved prompts, effectively bridging the knowledge gap for non-expert users. Our evaluation demonstrates \textit{PromptTailor}'s advantages across various methods and settings, highlighting particularly strong improvements for weaker models. The findings show that a lightweight student model, guided by powerful teachers, can effectively improve the open-ended responses of other LLMs, even though the optimizer itself is considerably smaller.

Future work could explore non-linear synthetic data generation with a more complex, multi-component prompt generation structure. While our current dialogue flow structure is linear for convenient fine-tuning, training on a more complex dataset could benefit from utilizing a Mixture-of-Experts (MoE) LoRA framework. This framework would allow individual LoRA adapters to be trained on specific dialogue components and activated dynamically during inference. 

\section*{Limitations}
Because of our limited scope and budget, we did not use the most powerful model as the backbone of our evaluator, and we recruited only ten participants for the human assessments. Additionally, while our LLM-based evaluator performs reliable pairwise comparisons, it lacks the ability to quantify response superiority. A marginally better response is currently treated as an absolute win. Future work should combine binary pairwise judgments with continuous metrics—such as average score—to better capture response quality differentials. For our experiment, we only compared our method (\textit{PromptTailor}) with a few other prompt optimization methods, such as CoT, ExpertPrompting, Evoke. However, most recent prompting methods only consider prompt optimization rather than generation tailored to user preferences, which increases the difficulty of direct comparison. We leave this as an open question for future work.

\section*{Ethical Considerations} 
Our human assessments were approved by the institution’s IRB. Each participant signed an informed-consent form and received a \$10 gift card as compensation. All participant information has been anonymized.

\bibliography{custom}

\appendix

\section{Appendix}
\label{sec:appendix}

\subsection{Domains and Themes for the Dataset}
\label{sec:domains}
\begin{enumerate}
\item Role-Play and Simulation -- characters, worlds, scenarios
\item Marketing, Brand and Social Strategy -- positioning, campaigns, community
\item Entertainment, Media and Pop Culture -- industry trends, fandoms, critique
\item Politics and Public Policy -- domestic governance, policy analysis
\item Sports and Athletics -- rules, tactics, commentary
\item Education and Instructional Design -- curricula, pedagogy, assessment
\item History and Historiography -- events, sources, methods
\item Science Communication -- explainers for lay audiences
\item Philosophy and Ethics -- logic, moral frameworks, dilemmas
\item Psychology and Behavioral Science -- cognition, behavior, motivation
\item Business Strategy and Leadership -- org models, scaling, executive practice
\item Career Coaching and Job Search -- resumes, interviews, growth plans
\item Health and Medicine -- conditions, care models, health systems
\item Software and Digital Transformation -- development topics, IT modernization
\item Environment and Sustainability -- climate, conservation, stewardship
\item Economics and Markets -- theory, policy, market dynamics
\item Law and Legal Literacy -- rights, procedures, legal concepts
\item Human Resources and People Ops -- hiring, performance, culture
\item Project and Program Management -- planning, delivery, PMO
\item Corporate Finance and Investing -- valuation, portfolio theory
\item Real Estate and Urban Development -- property, zoning, city growth
\item Travel and Tourism -- itineraries, cultural etiquette, destinations
\item Food and Culinary Arts -- techniques, cuisines, menu ideas
\item Fashion and Personal Style -- aesthetics, wardrobe systems
\item Creative Arts and Music Production -- visual design, composition, recording
\item Gaming and Interactive Media -- mechanics, design, esports
\item Relationships, Family and Parenting -- communication, development, dynamics
\item Personal Productivity and Time Management -- workflows, habits, tools
\item Home, DIY and Gardening -- projects, materials, horticulture
\item Pet Care and Animal Behavior -- training, enrichment, health basics
\item Automotive and Mobility -- vehicles, maintenance, transport tech
\item Fitness and Nutrition -- training programs, diet planning, performance
\item Mental Health and Wellbeing -- coping skills, mindfulness, supports
\item Spirituality and Comparative Religion -- beliefs, practices, traditions
\item Social and Cultural Studies -- institutions, norms, anthropology
\item International Relations and Geopolitics -- states, alliances, strategy
\item Journalism and Media Literacy -- reporting, verification, bias detection
\item Communication, Persuasion and Rhetoric -- argumentation, framing, style
\item Artificial Intelligence and Data Science -- models, analytics, insights
\item Innovation and Design Thinking -- discovery, prototyping, iteration
\item Entrepreneurship and Startups -- idea validation, traction, funding
\end{enumerate}

\subsection{Synthetic Data Example}
\label{sec:sydata}

\begin{itemize} 
    \item \textbf{User intent and preference:} I need to draft a comprehensive report summarizing the quarterly financial performance for the last fiscal quarter. The report should highlight key revenue streams, expenditures, and profit margins. Please include a section on market trends impacting our sales and provide actionable recommendations for improving profitability in the next quarter. I prefer a concise, data-driven response and wants the report to be easily digestible for executive leadership.

\item \textbf{Intent Detection and Prompt Generation:}
\begin{lstlisting}[language=json,backgroundcolor=\color{yellow!10},numbers=none]
{
    "purpose": "To draft a comprehensive report summarizing quarterly financial performance, highlighting revenue, expenditures, and profit margins, including market trends and actionable recommendations for profitability improvement.",
    "context": "The report is for executive leadership, needs to be concise and data-driven, and covers the last fiscal quarter.",
    "desired_outcome": "A well-structured, concise financial report that includes specific financial metrics, analysis of market trends, and practical recommendations, easily understandable by executives.",
    "capability_information": {
    "explicit_inferred_capabilities": [
        "Summarizing quarterly financial performance",
        "Highlighting key revenue streams",
        "Highlighting expenditures",
        "Highlighting profit margins",
        "Including a section on market trends impacting sales",
        "Providing actionable recommendations for improving profitability",
        "Generating concise responses",
        "Generating data-driven responses",
        "Creating easily digestible content for executive leadership"
      ],
      "task_required_capabilities": [
        "Understanding financial terminology and concepts",
        "Synthesizing quantitative data into clear summaries",
        "Identifying and explaining relevant market trends",
        "Formulating strategic, actionable recommendations",
        "Structuring a formal report",
        "Adhering to specific formatting and length constraints"
      ]
    },
    "agent_plan": "The agent will first process financial data (assumed to be provided or accessible), then structure the report into sections for financial metrics, market trends, and recommendations. It will ensure the report is concise, data-driven, and tailored for executive readability.",
    "initial_prompt": "Generate a concise financial report for executive leadership summarizing last quarter's performance. Include revenue, expenditures, profit margins, market trends, and recommendations for improving profitability."
}
\end{lstlisting}

\item \textbf{Prompt Optimization 
Suggestions:} 
\begin{lstlisting}[language=json,backgroundcolor=\color{yellow!10},numbers=none]
{    "summary": {
      "purpose": "To draft a comprehensive report summarizing quarterly financial performance, highlighting revenue, expenditures, and profit margins, including market trends and actionable recommendations for profitability improvement.",
      "context": "The report is for executive leadership, needs to be concise and data-driven, and covers the last fiscal quarter.",
      "desired_outcome": "A well-structured, concise financial report that includes specific financial metrics, analysis of market trends, and practical recommendations, easily understandable by executives."
    },
    "optimized_capabilities": [
      "Synthesizing and summarizing complex financial data (revenue, expenditures, profit margins)",
      "Understanding and applying relevant financial terminology and concepts",
      "Analyzing and explaining significant industry-specific market trends and their impact on financial performance",
      "Formulating strategic, actionable recommendations for improving profitability",
      "Structuring professional, concise, and data-driven reports tailored for executive audiences",
      "Adhering to specified output format, length, and formatting requirements"
    ],
    "plan_prompt_improvement": "The goal is to transform the simple prompt into a highly effective instruction set that guides the LLM agent to produce a superior financial report. This involves clearly defining the agent's role, ensuring precise terminology is used, providing essential background, adding any missing details, and strictly specifying the output format and length, while balancing intent alignment with optimal LLM performance through refined capabilities.",
    "optimization_suggestions": [
      {
        "suggestion_number": 1,
        "title": "Define an appropriate role for the agent",
        "description": "Clearly establish the LLM's persona, such as 'You are a seasoned Financial Analyst and Report Generator.' This sets expectations for the tone, depth, and expertise required for the report."
      },
      {
        "suggestion_number": 2,
        "title": "Use precise, domain-specific terminology",
        "description": "Enhance clarity by incorporating specific financial terms like 'net profit margin,' 'operational expenditures,' or 'revenue streams' and clearly defining periods such as 'fiscal quarter (Q1 2025)' to ensure accuracy and professionalism."
      },
      {
        "suggestion_number": 3,
        "title": "Provide any necessary extra background context",
        "description": "Emphasize the report's audience and strategic purpose, for example, 'This report is critical for executive leadership to inform strategic decision-making and quarterly planning.'"
      },
      {
        "suggestion_number": 4,
        "title": "Add missing details for completeness",
        "description": "Specify any assumptions about data availability (e.g., 'Assume financial data for Q1 2025 will be provided in a structured format') or what specific elements to include in 'market trends' (e.g., 'Identify supply chain disruptions or shifts in consumer demand')."
      },
      {
        "suggestion_number": 5,
        "title": "Specify the desired output format, length, and formatting",
        "description": "Clearly state structural requirements like 'Present the report in markdown with distinct sections and subheadings,' 'Limit the total length to approximately 750-1000 words,' and 'Ensure all numerical data is presented in a clear, summarized table format where appropriate.'"
      }
    ]
}
\end{lstlisting}

\item \textbf{Optimized prompt:} 
\begin{lstlisting}[language=json,backgroundcolor=\color{yellow!10},numbers=none]

{
    "optimized_prompt": "You are a **seasoned Financial Analyst and Report Generator**, highly adept at **synthesizing complex financial data**, **analyzing market trends**, and **formulating strategic, actionable recommendations**. Your core mission is to create a **comprehensive, concise, and data-driven quarterly financial performance report** for **executive leadership**, providing clear insights to inform critical strategic decision-making.\n\nFor the **last fiscal quarter (Q1 2025)**, prepare a report that thoroughly summarizes key financial metrics, including **revenue streams, operational and capital expenditures, and net profit margins**. Present these figures with utmost clarity, preferably within a **summary table** or under distinct, well-labeled headings to ensure immediate readability for executives.\n\nCrucially, identify and explain any **significant industry-specific market trends** that impacted our sales and overall financial performance during this period. Your analysis should highlight both positive and negative influences.\n\nFinally, based on your comprehensive financial and market analysis, provide **concrete, actionable recommendations** aimed at improving profitability and optimizing financial performance for the **next fiscal quarter (Q2 2025)**. Ensure these recommendations are pragmatic and demonstrate clear potential for impact.\n\n**Output Requirements:**\n* **Format:** The report must be presented in **markdown format**, utilizing **clear, descriptive headings and subheadings** (e.g., `# Financial Performance Overview`, `## Revenue Analysis`, `### Key Trends`). Use **bullet points** for lists and recommendations to enhance readability.\n* **Length:** Aim for a report approximately **750-1000 words** in length, balancing detail with executive-level conciseness.\n* **Tone:** Maintain a **professional, analytical, and objective tone** throughout, suitable for high-level corporate review.\n* **Data Handling:** Assume the necessary raw financial data for Q1 2025 will be provided to you separately. Your focus should be entirely on **analysis, interpretation, and recommendation generation**. **Do not fabricate or 'hallucinate' specific financial figures if they are not explicitly given; use placeholders or descriptive language instead if specific numbers are not available.**"
}
\end{lstlisting}
    
\end{itemize}

\subsection{Automated Assessments Detail}
\label{sec:responses}

\paragraph{Evaluation examples:}
    \begin{itemize} 
    \item \textbf{Task:} Draft an overview of the safety precautions and travel advisories for solo female travelers in Southeast Asia, with specific country recommendations.
    \item \textbf{Response 1:} 

     Solo Female Travel in Southeast Asia: Safety Precautions and Travel Advisories**\n\n Southeast Asia is a popular destination for solo female travelers, offering a unique blend of culture, history, and adventure. However, as with any travel destination, it's essential to take necessary safety precautions and stay informed about local conditions. Here's an overview of safety precautions and travel advisories for solo female travelers in Southeast Asia, along with specific country recommendations.\n\n**General Safety Precautions:**\n\n1. **Research, research, research**: Before traveling to any new destination, research the local culture, customs, and laws to avoid unintentionally offending anyone.\n2. **Be mindful of your surroundings**: Pay attention to your environment, especially at night, and avoid walking alone in dimly lit or deserted areas.\n3. **Keep valuables secure**: Lock your luggage, keep your passport and important documents safe, and avoid carrying large amounts of cash.\n4. **Trust your instincts**: If you feel uncomfortable or unsafe, trust your instincts and avoid the situation.\n5. **Stay in touch with friends and family**: Keep your loved ones informed about your itinerary and expected return times.\n6. **Learn basic phrases in the local language**: Knowing basic phrases like hello, thank you, and help can go a long way in communicating with locals.\n7. **Respect local customs**: Dress modestly, remove your shoes when entering temples or homes, and avoid public displays of affection.\n\n**Country-Specific Safety Precautions and Travel Advisories:**\n\n1. **Thailand**: Thailand is generally a safe country for solo female travelers, but take necessary precautions in crowded areas and night markets. Avoid traveling alone at night, and be mindful of your belongings.\n2. **Vietnam**: Vietnam is a popular destination for solo travelers, but petty theft and scams are common. Keep your valuables secure, and avoid giving money to beggars or street performers.\n3. **Cambodia**: Cambodia is a developing country with limited infrastructure. Be prepared for dusty roads, poor sanitation, and occasional power outages. Avoid traveling alone at night, and be respectful of local customs.\n4. **Laos**: Laos is a laid-back country with a relaxed atmosphere. However, be prepared for limited healthcare facilities and occasional power outages. Avoid traveling alone at night, and be respectful of local customs.\n5. **Indonesia**: Indonesia is a vast and diverse country with many tourist attractions. However, be prepared for occasional earthquakes, volcanic eruptions, and terrorist attacks. Avoid traveling alone at night, and be respectful of local customs.\n6. **Malaysia**: Malaysia is a multicultural country with a mix of Malay, Chinese, and Indian influences. However, be prepared for occasional flooding and traffic congestion. Avoid traveling alone at night, and be respectful of local customs.\n7. **Philippines**: The Philippines is a beautiful country with stunning beaches and vibrant culture. However, be prepared for occasional typhoons, flooding, and terrorist attacks. Avoid traveling alone at night, and be respectful of local customs.\n\n**Specific Country Recommendations:**\n\n1. **Thailand**: Chiang Mai, Bangkok, and Phuket are popular destinations for solo female travelers. Take a cooking class, visit the temples, and enjoy the night markets.\n2. **Vietnam**: Hanoi, Ho Chi Minh City, and Hoi An are must-visit destinations. Try the local cuisine, visit the ancient city of Hoi An, and take a motorbike tour.\n3. **Cambodia**: Siem Reap and Angkor Wat are a must-visit for any traveler. Take a guided tour of the temples, and visit the bustling markets.\n4. **Laos**: Luang Prabang and Vientiane are charming towns with a relaxed atmosphere. Take a boat trip along the Mekong River, and visit the ancient temples.\n5. **Indonesia**: Bali and Yogyakarta are popular destinations for solo female travelers. Take a yoga class, visit the temples, and enjoy the beautiful beaches.\n6. **Malaysia**: Kuala Lumpur and Penang are vibrant cities with a mix of Malay, Chinese, and Indian influences. Try the local cuisine, visit the Petronas Twin Towers, and take a stroll along the beach.\n7. **Philippines**: Boracay and Palawan are stunning destinations with beautiful beaches and crystal-clear waters. Take a snorkeling or diving trip, visit the Chocolate Hills, and enjoy the local cuisine.\n\n**Conclusion:**\n\n Southeast Asia is a wonderful destination for solo female travelers, offering a unique blend of culture, history, and adventure. By taking necessary safety precautions and staying informed about local conditions, you can have a safe and enjoyable trip. Remember to research, be mindful of your surroundings, and respect local customs. With the right preparation and attitude, you can have a fantastic time exploring this beautiful region.

    \item \textbf{Response 2:} 
    
    Comprehensive Travel Guide for Solo Female Travelers in Southeast Asia**\n\n As a solo female traveler, Southeast Asia is an exciting and rewarding destination. With its rich cultural heritage, beautiful landscapes, and vibrant cities, it's no wonder why this region is a popular choice for travelers. However, solo female travelers may face unique challenges and concerns while traveling in Southeast Asia, including safety risks, cultural differences, and language barriers. In this comprehensive travel guide, we will provide you with the information and advice you need to stay safe, navigate cultural differences, and have a memorable and enjoyable trip to Southeast Asia.\n\n**Country-Specific Information and Recommendations**\n\n This guide covers at least 5 countries in Southeast Asia, including Indonesia, Thailand, Vietnam, Malaysia, and the Philippines. Each country has its unique culture, customs, and challenges, and we will provide you with country-specific information and recommendations to help you navigate each destination safely and confidently.\n\n**Indonesia**\n\n* Safety Precautions: Indonesia is generally a safe country, but petty theft and scams are common in tourist areas. Be mindful of your belongings, and avoid carrying large amounts of cash.\n* Travel Advisories: The US Department of State advises travelers to exercise increased caution in Indonesia due to terrorism, crime, and natural disasters.\n* Country-Specific Recommendations:\n\t+ Visit the ancient temples of Borobudur and Prambanan in Yogyakarta.\n\t+ Explore the vibrant city of Jakarta and its many markets and shopping centers.\n\t+ Take a boat trip to the Gili Islands for snorkeling and diving.\n\n**Thailand**\n\n* Safety Precautions: Thailand is generally a safe country, but be mindful of your belongings and avoid traveling alone at night.\n* Travel Advisories: The US Department of State advises travelers to exercise increased caution in Thailand due to terrorism, crime, and civil unrest.\n* Country-Specific Recommendations:\n\t+ Visit the bustling streets of Bangkok and its many markets and shopping centers.\n\t+ Explore the ancient city of Chiang Mai and its many temples and cultural attractions.\n\t+ Take a boat trip to the islands of Koh Samui and Koh Phangan for snorkeling and diving.\n\n**Vietnam**\n\n* Safety Precautions: Vietnam is generally a safe country, but be mindful of your belongings and avoid traveling alone at night.\n* Travel Advisories: The US Department of State advises travelers to exercise increased caution in Vietnam due to terrorism, crime, and civil unrest.\n* Country-Specific Recommendations:\n\t+ Visit the ancient city of Hoi An and its many temples and cultural attractions.\n\t+ Explore the vibrant city of Ho Chi Minh City and its many markets and shopping centers.\n\t+ Take a boat trip to Ha Long Bay for snorkeling and diving.\n\n**Malaysia**\n\n* Safety Precautions: Malaysia is generally a safe country, but be mindful of your belongings and avoid traveling alone at night.\n* Travel Advisories: The US Department of State advises travelers to exercise increased caution in Malaysia due to terrorism, crime, and civil unrest.\n* Country-Specific Recommendations:\n\t+ Visit the vibrant city of Kuala Lumpur and its many markets and shopping centers.\n\t+ Explore the ancient city of Malacca and its many temples and cultural attractions.\n\t+ Take a boat trip to the islands of Langkawi and Penang for snorkeling and diving.\n\n**Philippines**\n\n* Safety Precautions: The Philippines is generally a safe country, but be mindful of your belongings and avoid traveling alone at night.\n* Travel Advisories: The US Department of State advises travelers to exercise increased caution in the Philippines due to terrorism, crime, and natural disasters.\n* Country-Specific Recommendations:\n\t+ Visit the ancient city of Intramuros in Manila and its many cultural attractions.\n\t+ Explore the vibrant city of Cebu and its many markets and shopping centers.\n\t+ Take a boat trip to the islands of Palawan and Bohol for snorkeling and diving.\n\n**Practical Tips and Recommendations**\n\n* Accommodation: Consider staying in guesthouses or hostels, which are often cheaper and more social than hotels.\n* Transportation: Use local transportation such as buses, taxis, and motorbikes, which are often cheaper and more convenient than renting a car.\n* Food: Try local street food and markets, which are often cheaper and more authentic than restaurants.\n* Cultural Norms: Be respectful of local customs and traditions, and avoid public displays of affection.\n\n**Safety Precautions and Travel Advisories**\n\n* Petty theft and scams are common in tourist areas, so be mindful of your belongings and avoid carrying large amounts of cash.\n* Avoid traveling alone at night, and use reputable taxi services or ride-sharing apps.\n* Check local weather forecasts and natural disaster warnings before traveling.\n* Register with your government's travel advisory service, such as the US Department of State's Smart Traveler Enrollment Program (STEP), to receive important safety and security updates about your destination.\n\n**Conclusion**\n\n Southeast Asia is a wonderful destination for solo female travelers, with its rich cultural heritage, beautiful landscapes, and vibrant cities. However, solo female travelers may face unique challenges and concerns while traveling in Southeast Asia, including safety risks, cultural differences, and language barriers. In this comprehensive travel guide, we have provided you with country-specific information and recommendations to help you navigate each destination safely and confidently. Remember to always exercise increased caution, be mindful of your belongings, and respect local customs and traditions. Happy travels!
\end{itemize}

\begin{table*}[h]
   \centering
  \begin{tabular}{lcccc}
    \hline
    \textbf{Model} & \textbf{Response Comparison} & \textbf{1 (First Better)} & \textbf{2 (Second Better)} & \textbf{0 (Same)} \\
    \hline
    \textbf{Llama3-8B} & \textbf{Original vs. Llama3(FT)} & 12 & 22 & 6 \\
    \hline
    \textbf{Mistral-7B} & \textbf{Original vs. Llama3(FT)} & 5 & 26 & 9 \\
    \hline
  \end{tabular}
   \caption{\label{tab:human}\textbf{Results of human assessments, sample size 40}. We are comparing responses from Llama3-8B and Mistral-7B when given the original user intents versus the optimized prompt. “Original” refers to the original user prompt and "Llama3(FT)" denotes using our fine-tuned Llama3 model (\textit{PromptTailor}) to revise the prompt.}
   \label{tab:2}
\end{table*}

\subsection{Human Assessments}
\label{sec:humana}

For the human evaluation, we randomly selected 8 original user prompts from the test set. We recruited 10 participants from our campus community, mostly undergraduate students. Each participant used a web interface to compare responses side-by-side. Participants were divided into two groups, with each group assigned to compare either original vs Llama3-8B or original vs Mistral-7B. Participants were instructed to select the better response or indicate if the responses were of equal quality. During the assessment, participants needed to consider two metrics:
1) Alignment with User Intent: how well the response aligns with user intents and correctly meets user needs.
2) Response Quality: this includes the usefulness of information, style, correctness, and so on.
Participants were asked to record a score for each metric (on a scale of 1-10) and then average these scores. The response with a higher average score was designated the "better," or "same" if the scores were identical.

The aggregated results are shown in Table~\ref{tab:human}. We observed greater variation among individual human assessments: while some participants preferred the responses generated from the optimized prompt, others found the responses similar. However, the aggregated results in Table~\ref{tab:human} generally align with the automated assessment, although they show a greater preference for 0 (“same”).  

\begin{figure*}[t]              
  \centering
  \includegraphics[width=\linewidth]{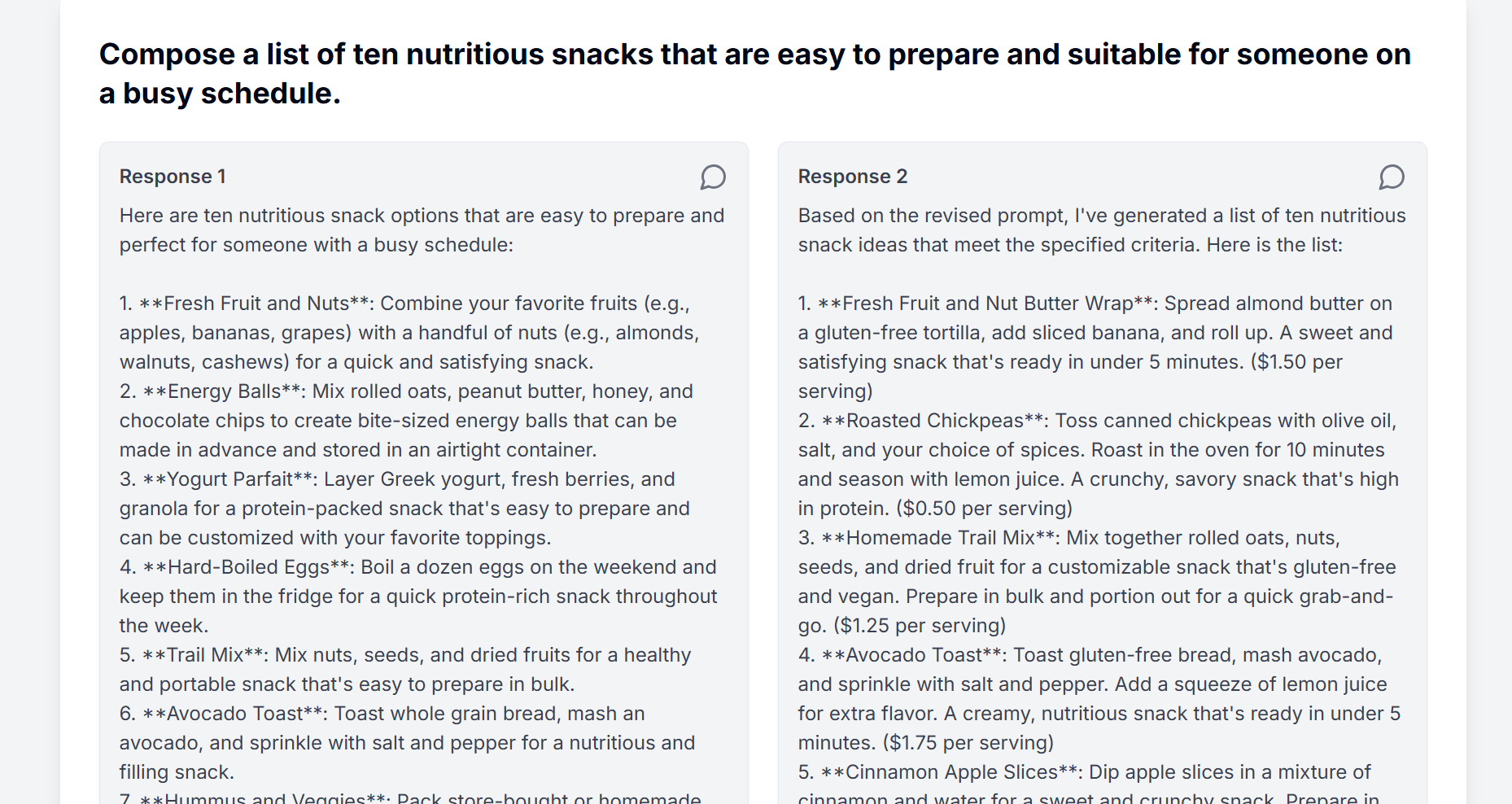}  
  \vspace{1cm}
  \includegraphics[width=\linewidth]{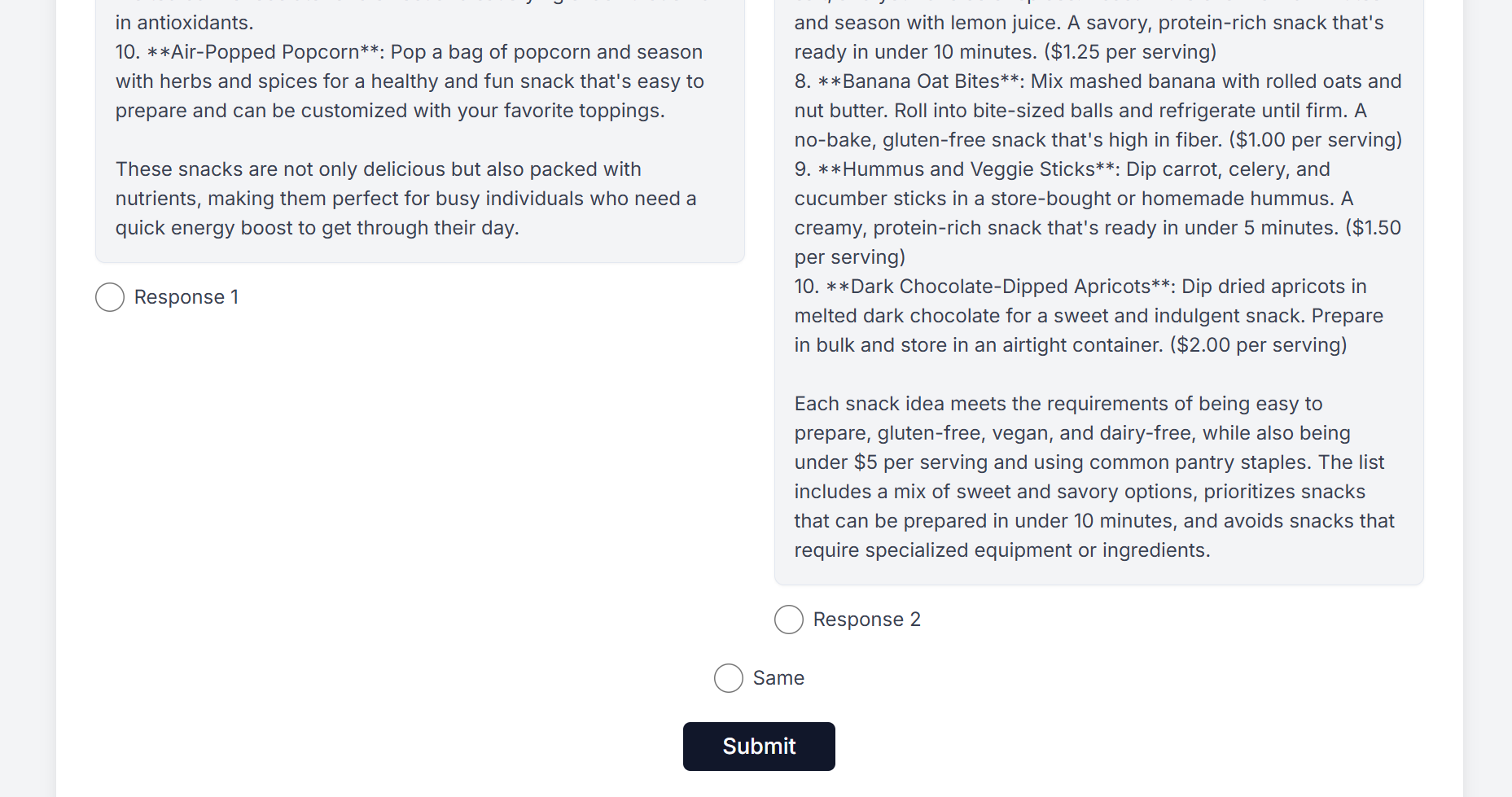}  
  \caption{Screenshots of our web interface.}
  \label{fig:web-ui}
\end{figure*}

\subsection{Human Assessments Web Interface}
\label{sec:human}
See Figure~\ref{fig:web-ui} for screenshots of the web interface.

\end{document}